\documentclass[pmlr]{jmlr}

\usepackage{longtable}

\usepackage[linewidth=1pt]{mdframed} 
 \usepackage{booktabs}
 
\usepackage[load-configurations=version-1]{siunitx} 

\makeatletter
\def\set@curr@file#1{\def\@curr@file{#1}} 
\makeatother

\theorembodyfont{\upshape}
\theoremheaderfont{\scshape}
\theorempostheader{:}
\theoremsep{\newline}

\newcommand{\website}{\url{http://ico-relations.ebm-nlp.com}}

\jmlrvolume{219}
\jmlryear{2023}
\jmlrworkshop{Machine Learning for Healthcare}

\title[Jointly Extracting Interventions, Outcomes, and Findings from RCT Reports with LLMs]{Jointly Extracting Interventions, Outcomes, and Findings from RCT Reports with LLMs}

\author{\Name{Somin Wadhwa}
       \Email{wadhwa.s@northeastern.edu}\\ 
       \addr Khoury College of Computer Sciences\\ Northeastern University\\
       Boston, MA, USA 
       \AND
       \Name{Jay DeYoung}
       \Email{deyoung.j@northeastern.edu}\\ 
       \addr Khoury College of Computer Sciences\\ Northeastern University\\
       Boston, MA, USA 
       \AND
        \Name{Benjamin Nye}
       \Email{bnye@coloradocollege.edu}\\ 
       \addr Mathematics \& Computer Science\\ 
       Colorado College\\
       Colorado Springs, CO, USA 
       \AND
       \Name{Silvio Amir}
       \Email{s.amir@northeastern.edu}\\ 
       \addr Khoury College of Computer Sciences\\ Northeastern University\\
       Boston, MA, USA 
       \AND
       \Name{Byron C. Wallace}
       \Email{b.wallace@northeastern.edu}\\ 
       \addr Khoury College of Computer Sciences\\ Northeastern University\\
       Boston, MA, USA }

\begin{document}

\maketitle


\begin{abstract}
  Results from Randomized Controlled Trials (RCTs) 
  establish the comparative effectiveness of interventions, and are in turn critical inputs for evidence-based care.
  However, results from RCTs are presented in (often unstructured) natural language articles describing the design, execution, and outcomes of trials; 
  clinicians must manually extract findings pertaining to interventions and outcomes of interest from such articles. 
  This onerous manual process has motivated work on (semi-)automating extraction of structured evidence from trial reports. 
  In this work we propose and evaluate a text-to-text model built on instruction-tuned Large Language Models (LLMs) to jointly extract 
  \textit{Interventions}, \textit{Outcomes}, and \textit{Comparators} (ICO elements) from clinical abstracts, and infer the associated results reported. 
  Manual (expert) and automated evaluations indicate that framing evidence extraction as a conditional generation task and fine-tuning LLMs for this purpose realizes considerable ($\sim$20 point absolute F1 score) gains over the previous SOTA.
  We perform ablations and error analyses to assess aspects that contribute to model performance, and to highlight potential directions for further improvements. 
  We apply our model to a collection of published RCTs through mid-2022, and release a searchable database of structured findings: \website.
\end{abstract}

\section{Introduction}
\label{section:intro}

Robust medical evidence concerning the comparative effectiveness of treatments is primarily disseminated in published free-text articles that report outcomes from randomized controlled trials (RCTs).
Such trial results are critical inputs for practicing \emph{Evidence-based medicine} (EBM; \citealt{sackett1997evidence}), which seeks to inform patient care using the totality of relevant findings.
Trial results are also potentially important for augmenting clinical predictions \citep{naik-etal-2022-literature}, and for calibrating trust in treatment suggestions offered by AI support systems \citep{yang2023harnessing}, which ought to agree with the established evidence.






\begin{figure}
    \centering
    \includegraphics[scale=0.6]{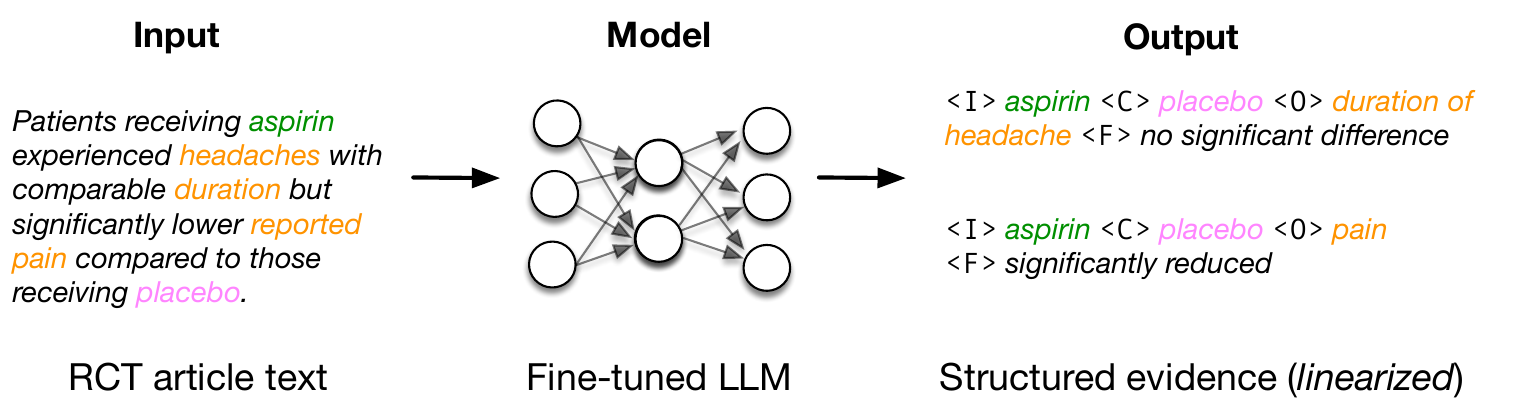}
    \caption{We fine-tune a Large Language Model (LLM) to map from free-text descriptions of clinical trials to structured representations of findings.}
    \label{fig:the-task}
\end{figure}

A challenge to making use of all available evidence is that findings from trials are disseminated via unstructured published articles. 
Researchers and healthcare providers must trawl through these to extract findings relevant to their clinical question(s). 
This problem has been exacerbated by the rapid production of new evidence: A now outdated estimate suggests that 75 trial reports are published \emph{every single day} \citep{bastian2010seventy}; more recent estimates put this number at $\sim$140 trial reports per day \citep{marshall2020trialstreamer}. 

To allow practitioners to draw upon newly published evidence as it accumulates, we need tools that make navigating findings more efficient. 
This has motivated work on Natural Language Processing (NLP) methods to semi-automate aspects of data extraction from clinical trial reports (\citealt{kang2021neuro,Kiritchenko2010,wallace2016extracting,nye2022understanding}, \emph{inter alia}). 
In this work we capitalize on and extend recent advances in NLP, specifically \emph{instruction-tuned} LLM capabilities \citep{chung2022scaling}, to perform end-to-end structured evidence extraction from free-text (Figure \ref{fig:the-task}).
We achieve state-of-the-art (SOTA) performance on this challenging task: The model we introduce yields a $\sim$20 point absolute gain in F1 score over the prior SOTA approach.
We ablate model components to assess their contributions.
We also release model weights, and a database of structured findings inferred by our model over a comprehensive dataset of articles describing RCTs. 

\subsection*{Generalizable Insights about Machine Learning in the Context of Healthcare}

With respect to \emph{healthcare}, this work makes significant progress on the important practical problem of structured evidence extraction from published articles describing RCTs.
The outputs of this system may aid evidence synthesis, and might also serve as inputs to other machine learning models in healthcare which could benefit from conditioning on robust evidence.
Beyond this, the need for data extraction from free-text (e.g., clinical notes) is widespread in healthcare: Improved extraction methods have the potential to ultimately allow clinicians to focus on providing patient care instead of navigating unstructured data. 

In terms of \emph{machine learning}, we introduce and evaluate a method for training LLMs to perform a complex instance of \emph{relation extraction}, a long-standing problem in ML \citep{ireson2005evaluating}. 
To our knowledge, this is one of the first efforts to evaluate LLMs for medical relation extraction; we find that they outperform existing systems for this task by a large margin. 
As an additional contribution which may be of interest to the broader machine learning community, our ablations indicate that including \emph{evidence spans} in extraction targets is an an important design decision---this complements recent developments inducing LLMs to provide free-text ``rationales'' for their outputs \citep{wei2022chain}, and may have implications for those working with LLMs for relation extraction going forward.



\section{Related Work}
\label{section:related-work}

In this work we develop and evaluate methods using LLMs to extract results from clinical trial reports.
Information and Relation Extraction (RE), generally, are well established sub-fields within NLP 
\citep{cowie1996information}, and we do not attempt to provide a general survey here. 
Instead, 
we contextualize our work by reviewing closely related efforts that focus on:
(i) Information extraction from biomedical/clinical texts (Section \ref{section:related-ie-cinica}); (ii) Models for jointly identifying entities and inferring relations between them (Section \ref{section:joint-re}); and (iii) Recent approaches that treat 
RE 
as a \emph{text-to-text} problem, a strategy that we adopt here (Section \ref{section:generative-RE}).

\subsection{Information Extraction from Biomedical Literature and Clinical Text}
\label{section:related-ie-cinica}

A line of prior work in NLP attempts to extract relevant \textit{Populations, Interventions, Comparators} and \textit{Outcomes} (PICO elements) from clinical texts~\citep{kim2011automatic}. 
\cite{nye-etal-2018-corpus} collected a corpus of 5,000 annotated RCT abstracts and introduced novel NLP tasks aiding evidence-based medicine. \cite{10.1145/3331184.3331352} highlighted important aspects of PICO human-annotations to refine datasets by adopting a relaxed agreement schemes for human annotations of PICO. 
\cite{jin-szolovits-2018-pico} introduced baselines in detecting PICO elements at the sentence level using LSTMs. 
\cite{Schmidt2020DataMI} proposed framing PICO extraction as a question-answering task and subsequently using transformer models, including SciBERT \citep{beltagy-etal-2019-scibert} --- a masked language model pretrained on large-scale scientific data. 
These efforts either pre-dated Transformers, or used small encoder backbones, i.e., BERT \citep{devlin2018bert}, rather than the generative models we use here. 

Elsewhere, \cite{lehman-etal-2019-inferring} introduced the \emph{evidence inference} dataset 
which entailed inferring which medical treatments work with respect to a \emph{given} ICO-set of interest. 
Using this dataset as a starting point, \cite{nye2022understanding} considered the end-to-end task of extracting PICO elements 
\emph{and} inferring results (as opposed to performing inference for a given ICO triplet). 
They proposed an \textit{extractive} entity extraction-linking-inference (ELI) sequential approach for this challenging task, and showed that it yielded results superior to standard joint architectures for relation extraction \citep{wadden-etal-2019-entity}. 
We improve upon these earlier efforts by introducing an end-to-end \textit{generative} model for the task of medical evidence inference. 

\subsection{Jointly Extracting Entities and their Relations}
\label{section:joint-re}
Early work in RE used pipeline approaches 
comprising separate models to, first, extract entities from a span of text, and then infer relations between those entities (if any). 
More recently, researchers have introduced joint extraction models since they tend to reduce error propagation and can capitalize on the connections between 
entities and their relations \citep{wang-lu-2020-two}. 
Traditionally, such joint extraction methods principally worked by predicting ``BILOU'' tags (Beginning, Inside, Last, Outside, and Unit) for tokens in the input \citep{bekoulis-etal-2018-adversarial, Bekoulis_2018, miwa-bansal-2016-end, zheng-etal-2017-joint, verga-etal-2018-simultaneously}. 
Span-based approaches extend these methods 
by constructing spans of tokens and then labeling these 
with respect to specific entity types, which enables processing of 
overlapping entities~\citep{DBLP:journals/corr/abs-1909-07755, wadden-etal-2019-entity} . 

\subsection{Generative Relation Extraction}
\label{section:generative-RE}

Most earlier methods for identifying entities and extracting relations in free text trained models with a joint objective \citep{eberts-ulges-2021-end, wang-lu-2020-two}. 
The recent rise in (\textit{very}) large language models (LLMs) \citep{brown2020language, chung2022scaling} has motivated research into using these models for structured prediction tasks such as named entity recognition and RE ~\citep{nayak2019effective, paolini2021structured, huguet-cabot-navigli-2021-rebel-relation}. 
This usually entails \emph{linearizing}---that is, encoding into strings---the structured information and then tasking models with generating linearized target relations conditioned on corresponding inputs. 



Building on these efforts, we propose to train and evaluate models to conditionally \textit{generate} ICO spans, findings regarding the reported comparative effectiveness of the corresponding intervention compared to the comparator for the outcome in question, \emph{and supporting textual evidence}.
Specifically, we fine-tune an LLM to generate sets of linearized outputs (tuples) containing all the entities, relations, and supporting evidence from a given input RCT abstract (Figure \ref{fig:illustration}).

\begin{figure*}
\centering
  \includegraphics[scale=0.85]{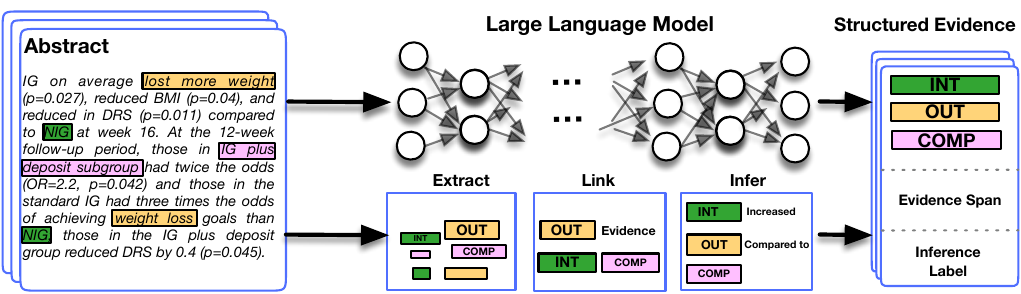}
  \caption{We propose instructional fine-tuning a large language model (top) using standard supervision to elicit evidence within generated ICO tuples. This approach yields substantial improvements over existing joint extraction approaches (bottom) where the entire task is decomposed into different \textit{independent} phases. }
  \label{fig:mainfig}
\end{figure*}

\section{Methods}
\label{section:methods} 

\subsection{End-to-End Evidence Inference}

The task of \textit{clinical evidence inference} 
comprises two sub-tasks: (i) Extraction of sets of relevant medical elements, i.e. ICO triplets;
and (ii) Inference regarding the effect of the primary intervention on the outcome (i.e., \textit{significant increase, significant decrease, no significant effect}), given the available evidence.
These two subtasks can be seen as specialized instances of entity tagging and relation extraction, respectively. 
Recent work on clinical evidence inference has adopted a sequential (pipeline) approach in which ICO extraction is treated as a sequence tagging step, and then a separate inference module processes the 
tagged entities \citep{nye2022understanding}.
This specialized approach outperformed model variants that attempted to jointly perform the task.
However, prior methods for joint extraction and inference pre-dated the modern LLMs which are the current dominant paradigm in NLP. 
Here we adopt such models, and treat the task of end-to-end evidence inference as a conditional language
generation task (Figure \ref{fig:mainfig}). 

Our targets are linearized strings comprising \textit{multiple} tuples, each containing the elements ({\emph{Intervention}, \emph{Comparator}, \emph{Outcome}, \emph{Evidence}, \emph{Inference label})}, extracted directly from an input abstract describing a RCT. 
Formally, given a RCT abstract $\mathcal{C}$, we model the probability of generating a linearized string $y$ of length $T$ containing $N$ tuples (separated by special tokens in the linearized forms), conditioned on $\mathcal{C}$:



\begin{equation*}
    p_{\text{LM}}(y | \mathcal{C}) = \prod_{t=1}^{T}p(y_t | \mathcal{C}, y_{<t})
\end{equation*}


\noindent This is the standard (conditional) language modeling objective, and we optimize for per token cross-entropy loss.
During training, we ``teacher force'', i.e., condition production of target token $y_t$ on the reference sequence $y_{<t}$ and $\mathcal{C}$.
At test time, the model iteratively conditions on its own outputs (we use greedy decoding).

The number of tuples associated with inputs is variable; language model flexibly models this by allowing the model to produce a special {\tt EOS} token after enumerating all tuples.
Note, however, that the model is unconstrained, and so can---and sometimes does, as we discussion in Section \ref{sec:errors}---produce invalid outputs (i.e., which do not conform to the linearized structured we assume).

\begin{figure*}
\centering
  \includegraphics[scale=0.48]{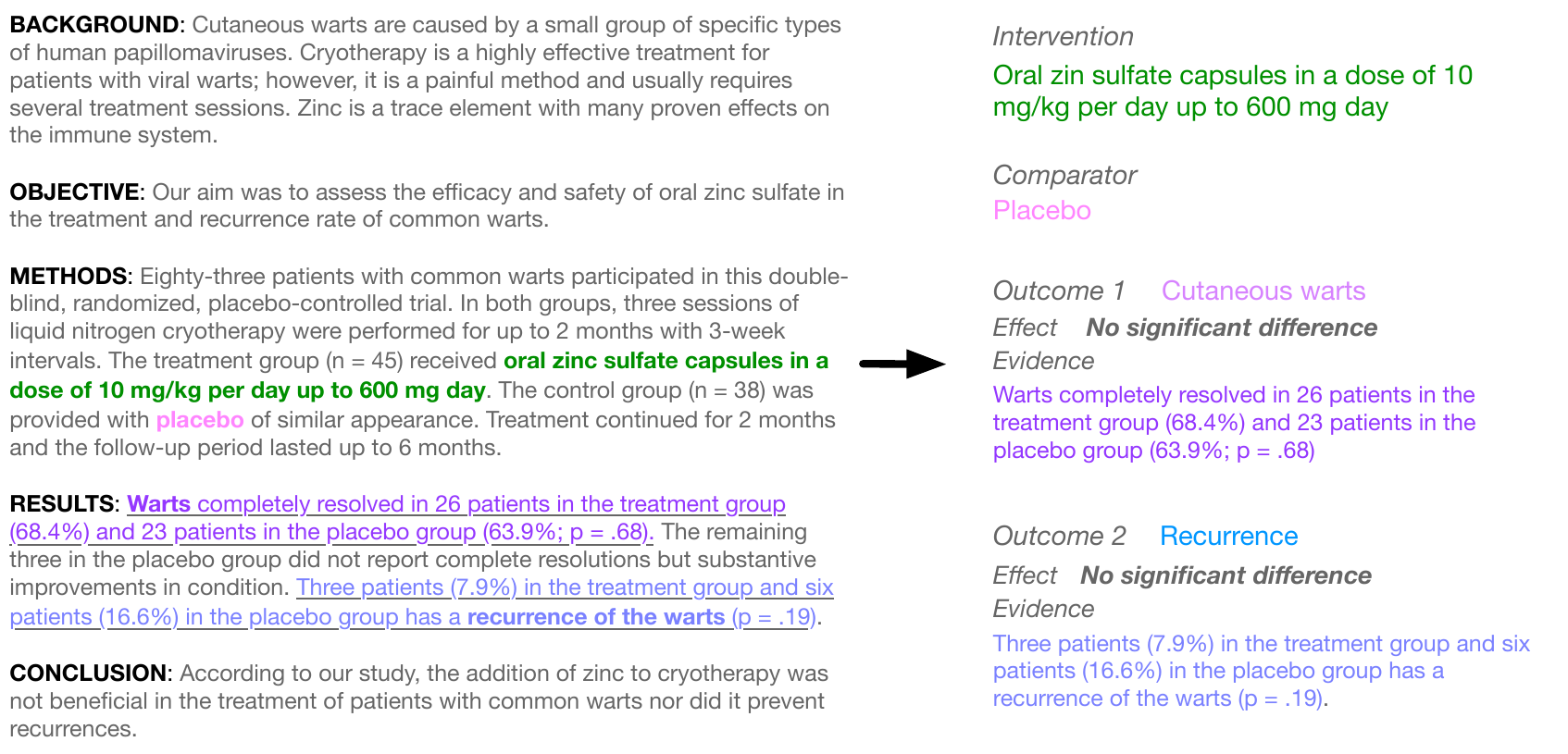}
  \caption{An illustration of the full evidence inference task. An end-to-end model is expected to extract all ICOs for which results were reported (highlighted here in pink, green, and orange) in an abstract describing an RCT, 
  and infer a label (\emph{significant increase}, \emph{significant decrease}, \emph{no significant difference}) based on the relevant evidence snippets which are also to be output (underlined here).}
  \label{fig:illustration}
\end{figure*}

Figure \ref{fig:illustration} provides an illustrative example where the abstract comprises  two unique reference tuples: 
\begin{flushleft}
    \texttt{(zinc sulfate capsules, placebo, warts, \textit{warts resolved in $68\%$ of the patients in treatment group and $64\%$ of the patients in placebo group}, \textbf{no significant difference})}
\end{flushleft}
\begin{flushleft}
    \texttt{(zinc sulfate capsules, placebo, recurrence of warts, \textit{three patients in treatment group and six patients in placebo group had a recurrence of warts (p=$.19$)}, \textbf{no significant difference})}
\end{flushleft}




\subsection{Data}
We derived the data we use for training from the Evidence Inference dataset \citep{lehman-etal-2019-inferring,deyoung-etal-2020-evidence}.
This comprises articles describing RCTs annotated by medical doctors.\footnote{Although the full dataset contains full-text RCT reports, here use use an abstract-only subset.} 
An instance in this dataset comprises an abstract annotated with five elements: An ICO triplet,
a \textit{label} that indicates the directionality of a reported effect of the intervention for the given outcome relative to the comparator (i.e., categorizing that the intervention yielded \textit{statistically significant increase, decrease, no effect} with respect to the outcome), 
and 
an \textit{evidence snippet}. 
The latter is an excerpt from the abstract providing support for a particular label.
This may be viewed as an explanation or ``rationale''. 
Together, these five elements form our targets. 
Table \ref{tab:datasets} provides basic data statistics for our training, validation, and test sets. 

\paragraph{Evaluation Data} To get an accurate assessment of model performance, \cite{nye2022understanding} also collected \textit{exhaustive} manual annotations from medical experts for 160 RCT abstracts. 
Owing to the inherent noise in distantly-supervised training lables, we observed that human annotators often identify substantially more tuples per abstract --- $4.97$ tuples per abstract in the \textit{validation} set, and $4.01$ in the \textit{test} set, as opposed to $2.76$ in the (non-exhaustive) \textit{training} set (Table \ref{tab:datasets}). We provide more detailed examples of this phenomenon in our error analysis in Section \ref{sec:errors}. 


\begin{table}[]
\centering
\begin{tabular}{lrrrrrr}
\hline
                    & \multicolumn{2}{c}{Train} & \multicolumn{2}{c}{Dev} & \multicolumn{2}{c}{Test} \\ \hline
Abstracts           & 1,964       & (1.00)      & 46        & (1.00)      & 89        & (1.00)       \\
Total ICO Tuples    & 5,430       & (2.76)      & 229       & (4.97)      & 357       & (4.01)       \\
Unique ICO Triplets & 4,951       & (2.52)      & 224       & (4.86)      & 351       & (3.94)      
\end{tabular}
\caption{Dataset statistics. We report the number of abstracts and the number of relations per abstract (denoted parenthetically). Development and test set statistics differ from their source \citep{nye2022understanding} as we omit documents with no annotated relations. 
}
\label{tab:datasets}
\end{table}

\subsection{Experimental Setup}
We performed all of our experiments on a single NVIDIA Quadro RTX 8000 GPU. 
We used the Huggingface library (v4.26.1; \citealt{wolf-etal-2020-transformers}) and publicly available checkpoints.\footnote{\url{https://huggingface.co/docs/transformers/model_doc/flan-t5}} of the language models we used in our experiments 
Our best performing model was trained for 8 epochs with a learning rate of $1e-6$, batch size of 2 (for both training and evaluation), with a maximum input 
length of 1024, and maximum output length of 512. For hyperparameter tuning, we only varied the learning rate, and max epochs. The remaining hyperparameters were left to their default values.
We used the Adam optimizer without gradient accumulation or gradient checkpointing. 



\begin{table}[]
\centering
\begin{tabular}{@{}lrrr@{}}
\toprule
\textbf{Full Inference End to End}    & \multicolumn{1}{c}{Precision} & \multicolumn{1}{c}{Recall} & \multicolumn{1}{c}{F-1} \\ \midrule
BRAN \citep{verga-etal-2018-simultaneously}                      & 0.05                          & 0.41                       & 0.08                    \\
DyGIE++ \citep{wadden-etal-2019-entity}                 & 0.24                          & 0.13                       & 0.17                    \\
ELI \citep{nye2022understanding}                       & 0.33                          & 0.31                       & 0.32                    \\
\midrule
\multicolumn{4}{c}{\small{\textit{(end-to-end  generation of ICO triplets with labels and supporting evidence)}}}  \\
BART  \citep{lewis-etal-2020-bart}                     & 0.38                            & 0.33                         & 0.35                      \\ 
T5-\textit{base} \citep{raffel2020exploring}                   & 0.56                            & 0.35                         & 0.43                      \\
Flan-T5-\textit{base} \citep{chung2022scaling}              & 0.69                        & 0.43                     & 0.53                  \\
\textbf{Flan-T5-\textit{large}}              & \textbf{0.75}                        & \textbf{0.48}                     & \textbf{0.59}                  \\
Flan-T5-\textit{large} (without evidence span extraction) & 0.49                        & 0.36                     & 0.41                  \\ \bottomrule
\end{tabular}
\caption{End-to-end relation extraction results, compare to \citet{nye2022understanding} Table 2a}
\label{tab:res1}
\end{table}


\section{Results} 
\label{section:results}

We perform both an end-to-end evaluation (Table \ref{tab:res1}) and ablate performance over ICO-triplet extractions only (Table \ref{tab:res2}), maintaining comparability to existing work \citep{nye2022understanding}. 
Section \ref{sec:evaluation_approach} contains details of our manual evaluation, and Section \ref{sec:errors} a detailed error analysis of model performance.

\subsection{Evaluation}  \label{sec:evaluation_approach}

Open-ended free text generation poses challenges to the evaluation of model outputs. 
Past work in the area, especially prior to LLMs, tended to perform a ``strict'' evaluation \citep{taille-etal-2020-lets} requiring exact matches of entities and their corresponding relations to reference targets.
This was appropriate because the models were effectively annotating input tokens, and references are assumed to be extractive. 
By contrast, because they are abstractive, LLMs 
can produce a variety of outputs that convey the desired semantic content---i.e., aligned with the reference target---without matching words exactly.

\begin{table}[]
\centering
\begin{tabular}{@{}lrrr@{}}
\toprule
\textbf{ICO-Triplet Extraction} & \multicolumn{1}{c}{Precision} & \multicolumn{1}{c}{Recall} & \multicolumn{1}{c}{F-1} \\ \midrule
DyGIE++ \citep{wadden-etal-2019-entity}                         & 0.45                          & 0.47                       & 0.46                    \\
ELI  \citep{nye2022understanding}                            & 0.46                          & 0.69                       & 0.55                    \\ \midrule
\multicolumn{4}{c}{\textit{(end-to-end generation of ICO-triplets)}}                                                     \\
T5-base \citep{raffel2020exploring}                         & 0.68                          & 0.62                       & 0.65                    \\
Flan-T5-\textit{base} \citep{chung2022scaling}                     & 0.78                          & 0.68                       & 0.73                    \\
\textbf{Flan-T5-\textit{large} }                  & \textbf{0.85}                          & \textbf{0.74}                       & \textbf{0.79}                    \\ \bottomrule
\end{tabular}
\caption{ICO-Triplet Ablation, compare to \citet{nye2022understanding} Table 2b (entity extraction)}
\label{tab:res2}
\end{table}

This motivates manual evaluation of RE outputs.
Specifically, we recruited three medical doctors (domain experts) via the Upwork platform.\footnote{\url{https://upwork.com}. We paid these experts \$30/hour to evaluate generated tuples.}
We asked these experts to individually evaluate each reference (to measure precision) \textit{and} generated tuple (to measure recall) from our exhaustive test set. 
For each reference tuple we asked experts to indicate: (1) Whether the reference ICO triplet appears in the set of generated tuples for that given abstract; and (2) Whether the target tuple as a whole could be derived 
from the set of generated tuples for that given abstract. 
Similarly, for each generated tuple we asked annotators to indicate: (1) Whether the ICO triplet appears in the abstract; and (2) Whether the tuple as a whole is correct (i.e., if it also gets the relevant supporting evidence and reported directionality). 
We provide examples of each category in the Appendix A. 
Human evaluators achieved strong annotation agreement; Fleiss kappa, $\kappa=0.77$. All three evaluators chose the same relevance label $\sim$92.4\% of the time. 
We derived final (consensus) labels by simple majority vote. 

\subsection{Error Analysis} 
\label{sec:errors}

We now describe, and provide examples of, some of the recurring error types from our best performing model (Flan-T5-large) on the validation data, and a set of abstracts from approximately 660,000 RCTs from the Trialstreamer database.\footnote{\url{https://trialstreamer.ieai.robotreviewer.net/}}


\paragraph{Incorrectly structured outputs} The model sometimes generated incorrectly formatted outputs which cannot be evaluated because they do not conform to the expected structure. 
(Recall that the model is not explicitly constrained to yield outputs that follow the desired linearization scheme.)
These include generations where: (1) there are missing elements in the (partial) ICO triplets; (2) outputs have an invalid syntactic structure (and are thus unparseable by any downstream tools); (3) some elements are duplicated; (4) the output contains irrelevant or unrelated tokens. 
The following is an example of one such instance:

\begin{flushleft}
    \textbf{Generated}: \texttt{[none, score, no, none, score was not significantly different between the two groups., no significant difference]}
\end{flushleft}

\noindent Here the instance has an incorrect number of tuple elements ($6$ instead of $5$), multiple elements are invalid, and while it does produce a valid label (``no significant difference''), there are no primary intervention and outcome spans associated with the label. 
This behavior occurs 
in only a small fraction ($\sim$0.53\%) of the RCT abstracts from Trialstreamer we ran through our model. 

\paragraph{Opposite inference labels for same ICOs} Approximately $12.3\%$ of generated tuples had ICO-triplet matches in the reference set (i.e., the ICO triplet was correctly extracted), but the inferred label regarding the reported findings concerning these was incorrect (e.g., significant \textit{increase} instead of significant \textit{decrease}). 
On inspection we found that such tuples belonged to two categories: (1) The primary intervention and comparator were swapped (leading to a flipped, albeit still correct, inference label with the same extracted evidence span); (2) Minor differences in generated \textit{outcomes} which resulted in a change in the label. 
The following is an example of the latter from our development set (PMID: $24227660$:\footnote{\url{https://pubmed.ncbi.nlm.nih.gov/24227660/}})

\begin{flushleft}
    \textbf{Abstract snippet}: \texttt{Canagliflozin increased urinary glucose excretion in a dose-dependent manner and produced statistically significant reductions in body weight compared with placebo (least squares mean percent changes from baseline of -2.2\%, -2.9\%, -2.7\%, and -1.3\% with canagliflozin 50, 100, and 300 mg and placebo; P $<$ 0.05 for all comparisons). Overall adverse event (AE) rates were similar across groups. Canagliflozin was associated with higher rates of genital mycotic infections in women, which were generally mild and led to few study discontinuations. Osmotic diuresis-related AE rates were low and similar across groups.}
\end{flushleft}

\begin{flushleft}
    \textbf{Reference}: \texttt{[canagliflozin, body weight, placebo, Canagliflozin increased urinary glucose excretion in a dose-dependent manner and produced statistically significant reductions in body weight compared with placebo., canagliflozin [LABEL] significantly decreased [OUT] body weight [COMP] placebo]}
\end{flushleft}

\begin{flushleft}
    \textbf{Generated}: \texttt{[canagliflozin, body weight reduction, placebo, Canagliflozin increased urinary glucose excretion in a dose-dependent manner and produced statistically significant reductions in body weight compared with placebo., canagliflozin [LABEL] significantly increased [OUT] body weight reduction [COMP] placebo]}
\end{flushleft}

\noindent An \textit{increase} in \textit{body weight reduction} is functionally the same as a \textit{decrease} in \textit{body weight}, and this explains the label flip. 

\paragraph{Combining multiple tuples} On average, our best performing model generates 3.49 ICO tuples per instance, as opposed to 4.01 per instance in the reference test set (Table \ref{tab:datasets}). 
This difference appears to be due to the model \textit{combining} multiple interventions and/or outcomes into one in cases where the inference label is preserved, in turn reducing the number of generated tuples. 
Consider the following example\footnote{Example simplified for brevity.} from our dev set where this behavior can be observed (PMID: 27981024\footnote{\url{https://pubmed.ncbi.nlm.nih.gov/27981024/}}):

\begin{flushleft}
    \textbf{Reference}: \texttt{[memory game with fruit, banana intake, no fruit game,  evidence, significant increase], \quad [memory game with fruit, mandarin intake, no fruit game, evidence, significant increase]}
\end{flushleft}

\begin{flushleft}
    \textbf{Generated}: \texttt{[fruit version of memory game, intake of mandarins and bananas, no fruit game,  evidence, significant increase]}
\end{flushleft}

\noindent Here we can observe that the generated tuple has combined banana and mandarin intake, yielding a single output instead of the two in the reference. 

\paragraph{Correctly generated but without any corresponding reference} This  type of ``error'' is limited to non-exhaustive reference sets, and occurs when there is no corresponding reference tuple for a correctly generated ICO output (because the reference set is non-exhaustive). 
While this is rare, instances featuring this type of error highlight the utility of the retrieved evidence snippets, which can be used to verify the output.
The following is one such example taken from an abstract that was \textbf{not} exhaustively annotated (PMID: 28211020\footnote{\url{https://pubmed.ncbi.nlm.nih.gov/28211020/}}):

\begin{flushleft}
    \textbf{Abstract snippet:}\footnote{Shortened for brevity.}   \texttt{High-risk patients undergoing brain surgery were randomly assigned to a usual care group (control group) or a GDFR group. In the GDFR group, (1) fluid maintenance was restricted to 3 ml/kg/h of a crystalloid solution and (2) colloid boluses were allowed only in case of hypotension associated with a low cardiac index and a high stroke volume variation. The primary outcome variable was ICU length of stay, and secondary outcomes were lactates at the end of surgery, postoperative complications, hospital length of stay, mortality at day 30, and costs. ICU length of stay was shorter (3 days [1-5] vs. 6 days [3-11], p $=$ 0.001) and ICU costs were lower in the GDFR group. The total number of complications (46 vs. 99, p = 0.043) and the proportion of patients who developed one or more complications (19.2 vs. 34.7\%, p $=$ 0.034) were smaller in the GDFR group. Hospital length of stay and costs, as well as mortality at 30 day, were not significantly reduced.}
\end{flushleft}

\begin{flushleft}
    \textbf{Reference}: \texttt{[goal-directed fluid restriction (GDFR) strategy, median hospital length of stay, usual care,  \textit{Hospital length of stay and costs, as well as mortality at 30-day, were not significantly reduced}, no significant difference]}
\end{flushleft}

\begin{flushleft}
    \textbf{Generated}: \texttt{[goal-directed fluid restriction (GDFR), ICU length of stay, usual care group (control group),  \textit{ICU length of stay was shorter (3 days [1–5] vs. 6 days [3–11], p = 0.001) and ICU costs were lower in the GDFR group.}, significantly decreased]}
\end{flushleft}

\noindent Here, on inspecting only the 
ICO-triplet and the inference label, one might assume that the generated tuple is incorrect with respect to the reference (due to a flipped label). 
However, examining the extracted 
evidence in support of the label and 
the full abstract reveals that the study does indeed report \textit{median length of hospital stay} and \textit{ICU-length of stay} as separate outcomes with different (opposite) labels.



\label{section:prototype} 

\begin{figure*}
\centering
  \includegraphics[scale=0.26]{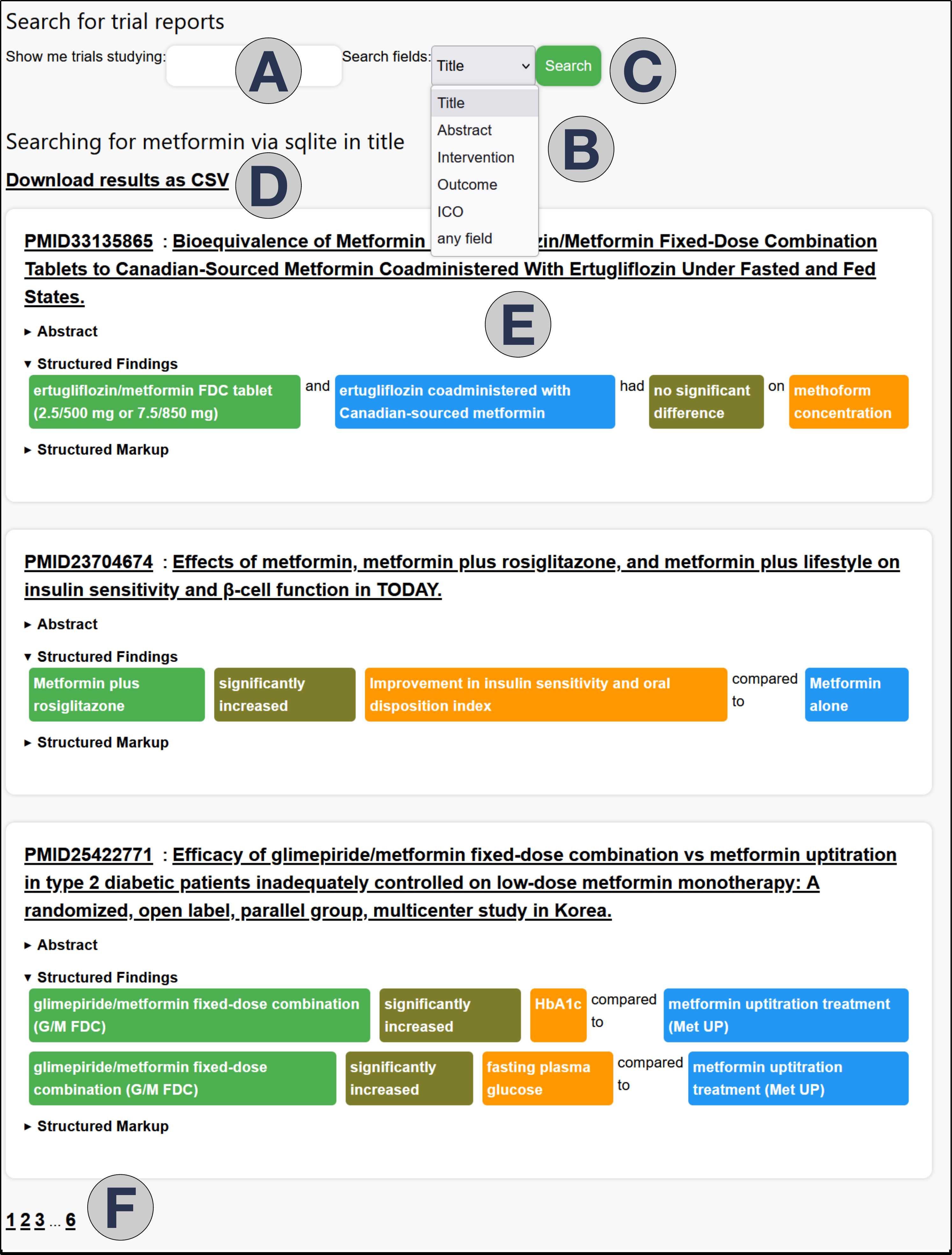}
  \caption{A screenshot of our prototype search interface over structured evidence. (A) User inputs a search query and select the fields (B) 
  to be searched over via an SQL search (C; \citealt{sqlite2020hipp}, e.g., entire abstract, only ICOs). 
  Search results can either be downloaded as a structured CSV (D) or the user can browse through individual results (E). We retrieve up to 100 documents per search query with 10 documents per page (F). The interface allows the users to read expanded abstracts, view structured findings \emph{(shown above)}, and expand structured markup for a tabular view of findings.} 
  \label{fig:interface}
\end{figure*}

\section{A Prototype for Browsing Structured Evidence}
To further demonstrate the (potential) utility of 
structured evidence extraction over the published evidence base, we make available a demonstration web application.\footnote{Hosted at \website.} 
This permits free-text search, 
which retrieves relevant structured evidence extracted from papers (we also link back to the original PubMed articles). 

We processed all Randomized Control Trials indexed by Trialstreamer \citep{marshall2020trialstreamer} as of June 2022, yielding 657,698 total studies and a total of 1,204,027 extracted relations. 
Relation extraction required 584 GPU (32GB NVIDIA V100) hours. 
Of the 770,356 unique Trialstreamer documents, approximately 50k instances were missing a full abstract. 
When processed via FLAN, 74k (about 10\%) had an unparseable output; lacking (or possessing an extra) a syntatic element (e.g. missing a bracket or having an extra one, or other terminator symbol). Another 5k had an output with an incorrect number of fields. 
82 had a malformed label. 
When parsing misclassified RCTs (erroneously included in Trialstreamer), the model would hallucinate ICOs and findings not present in the data. 


The prototype implements a BM25 search \citep{Robertson1994OkapiAT} backed by SQLite \citep{sqlite2020hipp}, allowing for search over multiple fields.\footnote{We experimented with embedding based methods but were ultimately disappointed with results}
The website allows for downloading search results (by search or by list of PMIDs/PMCIDs); our hope is that this may be of interest to researchers. 
We will make the entire raw database of inferred relations available upon publication. 

\section{Discussion} 


We have 
introduced and evaluated a state-of-the-art approach to 
end-to-end structured evidence extraction from natural language articles describing the conduct and results of clinical trials.
Specifically, we treat this problem as a conditional generation task and fine-tune Flan-T5 \citep{chung2022scaling}---a modestly sized instruction-tuned sequence-to-sequence model---to consume unstructured texts and yield structured tuples composed of interventions, comparators, outcomes and the results reported regarding these.
The latter comprises a discrete prediction encoding the direction of the reported finding, and a snippet of evidence supporting this determination. 
Ablations indicate the importance of jointly extracting evidence spans to support the inference task; this may have implications for work on relation extraction via conditional generative models more broadly.

Structured evidence extraction is an important task for realizing the promise of evidence-based medicine (EBM; \citealt{sackett1997evidence}), which aspires to inform treatment decisions on the basis of all available relevant evidence. 
The vast (unstructured) evidence base and rapid accumulation of new findings render practicing EBM challenging. 
The proposed approach to evidence extraction achieves substantially better performance than the prior state-of-the-art \citep{nye2022understanding}, and this brings us closer to being able to synthesize all evidence relevant to a given query, in real-time.

To illustrate the potential utility of this model, we have also made available a prototype interface that permits search directly over structured evidence tuples automatically extracted from a comprehensive database of randomized controlled trial reports. 
Our hope is that this demonstrates the precision of model outputs, and suggests how such extracted evidence might help researchers and healthcare providers navigate the evidence base more efficiently than is currently possible.
We also anticipate that the resultant database (comprising tuples from all RCTs in humans) may be a useful resource for researchers in machine learning for healthcare broadly, as one might draw upon such trial results to inform and/or justify ML predictions \citep{yang2023harnessing,naik-etal-2022-literature}.


\paragraph{Limitations} This work has several important limitations. 
First, while we have reported promising empirical results, the model we have trained here still makes errors (e.g., provides inexhaustive extractions from an inputs; see Section \ref{sec:errors}).
Any downstream use of the structured evidence outputs need to take this into account.

A methodological limitation is that we did not investigate the capabilities of even larger LLMs like GPT-3.5/4 \cite{brown2020language} for this task.
One could, in principle, use OpenAI's API to fine-tune such models for this task, and given their size it is likely that this would yield (probably moderately) improved results. 
We opted not to pursue this primarily because we prefer to use open-source models, to ensure scientific transparency and so that we can release model weights. 
Furthermore, the main contribution here is the framing of the task as a language modeling problem; the particular choice of underlying LLM is a secondary consideration. 

Finally, while we think structured evidence in the format that we have extracted---providing explicit sets of interventions, comparators, outcomes and evidence concerning these---will provide meaningful downstream utility for those interested in navigating and making sense of the published evidence base, it is currently an intermediate output.
The actual utility of this sort of model for downstream tasks which ultimately might affect care will require conducting further research.



\acks{This work was supported by the National Institutes of Health (NIH) under award R01LM012086, and by the National Science Foundation (NSF) award 1750978.}

\bibliography{anthology,custom}
\pagebreak
\appendix
\section*{Appendix A: Examples}
\label{appendix:examples}
Here we provide some full length output examples generated directly from our best performing models. Abstracts have been shortened for brevity to include key parts. 
\begin{mdframed}
\paragraph{PMID} $1457358$
\paragraph{Abstract}  Micronuclei reflect DNA damage in exfoliated cells and may thus provide a marker of early-stage carcinogenesis. Pre-treatment blood levels of cotinine, beta-carotene, retinol and vitamins C and E were similar in the placebo group (n = 61) and the treatment group (n = 53). Plasma beta-carotene levels increased 13-fold in the treatment group during intervention. Initial micronuclei counts (per 3,000 cells) were higher in the treatment group than in the placebo group (5.0 vs 4.0, P < 0.05). During intervention, the treatment group showed a 47$\%$ decrease, whereas the placebo group showed a non-significant decrease (16$\%$). After adjustment for the initial levels, the treatment group had 27$\%$ lower micronuclei counts than the placebo group at the end of the trial (95$\%$ CI: 9-41$\%$). These results indicate that beta-carotene may reduce lung cancer risk in man by preventing DNA damage in early-stage carcinogenesis.

\paragraph{Reference} [\textit{beta-carotene supplementation (20 mg d-1), Initial micronuclei counts (per 3,000 cells), placebo, Initial micro nuclei counts (per 3,000 cells) were higher in the treatment group than in the placebo group (5.0 vs 4.0, P; 0.05)., }\textbf{[INT]} \textit{beta-carotene supplementation (20 mg d-1)} \textbf{[LABEL]} \textit{significantly increased} \textbf{[OUT]} \textit{Initial micronuclei counts (per 3,000 cells)} \textbf{[COMP]} \textit{placebo}]

\paragraph{Generated} [\textit{14 weeks of beta-carotene supplementation (20 mg d-1), micronuclei counts, placebo, Initial micronuclei counts (per 3,000 cells) were higher in the treatment group than in the placebo group (5.0 vs 4.0, P  0.05).,} \textbf{[INT]} \textit{14 weeks of beta-carotene supplementation (20 mg d-1)} \textbf{[LABEL]} \textit{significantly increased} \textbf{[OUT]} \textit{micronuclei counts} \textbf{[COMP]} \textit{placebo}]
\end{mdframed}

\begin{mdframed}
    \paragraph{PMID}  $29295869$
    \paragraph{Abstract} Since sCD163 is shed to serum by inflammatory signals including lipopolysaccharides (LPS, endotoxin), we investigated sCD163 and correlations with lipid metabolism following LPS exposure. Eight healthy male subjects were investigated on two separate occasions: (i) following an LPS exposure and (ii) following saline exposure. Each study day consisted of a four-hour non-insulin-stimulated period followed by a two-hour hyperinsulinemic euglycemic clamp period. A 3H-palmitate tracer was used to calculate the rate of appearance (Rapalmitate). Blood samples were consecutively obtained throughout each study day. Abdominal subcutaneous adipose tissue was obtained for western blotting. We observed a significant two-fold increase in plasma sCD163 levels following LPS exposure (P < 0.001), and sCD163 concentrations correlated positively with the plasma concentration of free fatty acids, Rapalmitate, lipid oxidation rates and phosphorylation of the hormone-sensitive lipase at serine 660 in adipose tissue (P $<$ 0.05, all). Furthermore, sCD163 concentrations correlated positively with plasma concentrations of cortisol, glucagon, tumour necrosis factor (TNF)-$\alpha$, interleukin (IL)-6 and IL-10 (P $<$ 0.05, all). We observed a strong correlation between sCD163 and stimulation of lipolysis and fat oxidation following LPS exposure. These findings support preexisting theory that inflammation and macrophage activation play a significant role in lipid metabolic adaptions under conditions such as obesity, DM2 and NAFLD.

    \paragraph{Reference} [\textit{LPS exposure, macrophage activation, Saline axposure, We observed a significant two-fold increase in plasma sCD163 levels following LPS exposure (P $<$ 0.001), and sCD163 concentrations correlated positively with the plasma concentration of free fatty acids, Rapalmitate, lipid oxidation rates and phosphorylation of the hormone-sensitive lipase at serine 660 in adipose tissue (P $<$ 0.05, all). Furthermore, sCD163 concentrations correlated positively with plasma concentrations of cortisol, glucagon, tumour necrosis factor (TNF)-$\alpha$, interleukin (IL)-6 and IL-10 (P $<$ 0.05, all).}, \textbf{[INT]} \textit{LPS exposure} \textbf{[LABEL]} \textit{significantly increased} \textbf{[OUT]} \textit{macrophage activation} \textbf{[COMP]} \textit{Saline axposure}]

    \paragraph{Generated} [\textit{lipopolysaccharides (LPS) exposure, plasma sCD163 levels, saline exposure, We observed a significant two-fold increase in plasma sCD163 levels following LPS exposure (P $<$ 0.001), and sCD163 concentrations correlated positively with the plasma concentration of free fatty acids, Rapalmitate, lipid oxidation rates and phosphorylation of the hormone-sensitive lipase at serine 660 in adipose tissue (P $<$ 0.05, all).,} \textbf{[INT]} \textit{lipopolysaccharides (LPS) exposure} \textbf{[LABEL]} \textit{significantly increased} \textbf{[OUT]} \textit{plasma sCD163 levels} \textbf{[COMP]} \textit{saline exposure}]
\end{mdframed}

\begin{mdframed}
    \paragraph{PMID}  $26258157$
    \paragraph{Abstract} We conducted a double-blind, randomized trial of 134 outpatients with polymerase chain reaction-confirmed influenza to assess the effects of oseltamivir initiated 48–119 hours after illness onset. Oseltamivir treatment did not reduce illness duration, severity, or duration of virus detection. However, the power of this study was limited due to lower than expected enrollment.
    \paragraph{Reference 1} [\textit{Oseltamivir, Severity of illness, Control, Oseltamivir treatment did not reduce illness duration, severity, or duration of virus detection.,} \textbf{[INT]} \textit{Oseltamivir} \textbf{[LABEL]} \textit{no significant difference} \textbf{[OUT]} \textit{Severity of illness} \textbf{[COMP]} \textit{Control}]
    \paragraph{Reference 2} [\textit{Oseltamivir, Duration of virus detection, Control, Oseltamivir treatment did not reduce illness duration, severity, or duration of virus detection.,} \textbf{[INT]} \textit{Oseltamivir} \textbf{[LABEL]} \textit{no significant difference} \textbf{[OUT]} \textit{Duration of virus detection} \textbf{[COMP]} \textit{Control}]
    \paragraph{Generated} [\textit{Oseltamivir, Illness duration, severity, or duration of virus detection, Control, Oseltamivir treatment did not reduce illness duration, severity, or duration of virus detection.,} \textbf{[INT]} \textit{Oseltamivir} \textbf{[LABEL]} \textit{no significant difference} \textbf{[OUT]} \textit{Illness duration, severity, or duration of virus detection} \textbf{[COMP]} \textit{Control}]
\end{mdframed}

\begin{mdframed}
    \paragraph{PMID} $26283840$
    \paragraph{Abstract} Most commonly practiced surgical "lay open" technique to treat fistula-in-ano (a common anorectal pathology) has high rate of recurrence and anal incontinence. Alternatively, a nonsurgical cost efficient treatment with Ksharasutra (cotton Seton coated with Ayurvedic medicines) has minimal complications. In our study, we have tried to compare these two techniques. A prospective randomized control study was designed involving patients referred to the Department of General Surgery in RG Kar Medical College, Kolkata, India, from January 2010 to September 2011. Among 50 patients, 26 were in Ksharasutra and 24 were in fistulotomy group. 86\% patients were male and 54\% of the patients were in the fourth decade. About 74\% fistulas are inter-sphincteric and 26\% were of trans-sphincteric variety. Severe postoperative pain was more (7.7\% vs. 25\%) in fistulotomy group, while wound discharge was more associated with Ksharasutra group (15.3\% vs. 8.3\%). Wound scarring, bleeding, and infection rate were similar in both groups. Ksharasutra group took more time to heal (mean: 53 vs. 35.7 days, P $=$ 0.002) despite reduced disruption to their routine work (2.7 vs. 15.5 days work off, P $<$ 0.001). Interestingly, pain experienced was less in Ksharasutra group, there was no open wound in contrast to fistulotomy and it was significantly cost effective (Rupees 166 vs 464).

    \paragraph{Reference} [\textit{Ksharasutra (cotton Seton coated with Ayurvedic medicines), number of days ``off-work'', fistulotomy, Ksharasutra group took more time to heal (mean: 53 vs 35.7 days, P $=$ 0.002) despite reduced disruption to their routine work (2.7 vs. 15.5 days work off, P $<$ 0.001).,} \textbf{[INT]} \textit{Ksharasutra (cotton Seton coated with Ayurvedic medicines)} \textbf{[LABEL]} \textit{significantly decreased} \textbf{[OUT]} \textit{number of days ``off-work''} \textbf{[COMP]} \textit{fistulotomy}]

    \paragraph{Generated 1} [\textit{Ksharasutra (cotton Seton coated with Ayurvedic medicines), Severe postoperative pain, fistulotomy, Severe postoperative pain was more (7.7\% vs. 25\%) in fistulotomy group, while wound discharge was more associated with Ksharasutra group (15.3\% vs. 8.3\%).,} \textbf{[INT]} \textit{Ksharasutra (cotton Seton coated with Ayurvedic medicines)} \textbf{[LABEL]} \textit{significantly decreased} \textbf{[OUT]} \textit{Severe postoperative pain} \textbf{[COMP]} \textit{fistulotomy}]

    \paragraph{Generated 2} [\textit{Ksharasutra (cotton Seton coated with Ayurvedic medicines), Wound scarring, bleeding, and infection rate, fistulotomy, Wound scarring, bleeding, and infection rate were similar in both groups.,} \textbf{[INT]} \textit{Ksharasutra (cotton Seton coated with Ayurvedic medicines)} \textbf{[LABEL]} \textit{no significant difference} \textbf{[OUT]} \textit{Wound scarring, bleeding, and infection rate} \textbf{[COMP]} \textit{fistulotomy}]
\end{mdframed}

\end{document}